\pdfoutput=1

\documentclass[11pt,a4paper]{article}
\usepackage[hyperref]{emnlp2018}
\usepackage[utf8]{inputenc}
\usepackage{times}
\usepackage{latexsym}

\usepackage{url}
\usepackage{multirow}

\aclfinalcopy 

\title{Deep contextualized word representations for detecting sarcasm and irony}

\author{Suzana Ilić\textsuperscript{1},
	    Edison Marrese-Taylor\textsuperscript{2},
        Jorge A. Balazs\textsuperscript{2},
        Yutaka Matsuo\textsuperscript{2}\\
        University of Innsbruck, Austria\textsuperscript{1} \\
        {\tt suzana.ilic@student.uibk.ac.at} \\
        Graduate School of Engineering, The University of Tokyo, Japan\textsuperscript{2}\\
        {\tt \{emarrese,jorge,matsuo\}@weblab.t.u-tokyo.ac.jp}}

\date{}

\begin{document}
\maketitle

\begin{abstract}
Predicting context-dependent and non-literal utterances like sarcastic and ironic expressions still remains a challenging task in NLP, as it goes beyond linguistic patterns, encompassing common sense and shared knowledge as crucial components. To capture complex morpho-syntactic features that can usually serve as indicators for irony or sarcasm across dynamic contexts, we propose a model that uses character-level vector representations of words, based on ELMo. We test our model on 7 different datasets derived from 3 different data sources, providing state-of-the-art performance in 6 of them, and otherwise offering competitive results. 
\end{abstract}

\section{Introduction}

Sarcastic and ironic expressions are prevalent in social media and, due to the tendency to invert polarity, play an important role in the context of opinion mining, emotion recognition and sentiment analysis \cite{Pang2006}. Sarcasm and irony are two closely related linguistic phenomena, with the concept of meaning the opposite of what is literally expressed at its core. There is no consensus in academic research on the formal definition, both terms are non-static, depending on different factors such as context, domain and even region in some cases \cite{Filatova2012}.

In light of the general complexity of natural language, this presents a range of challenges, from the initial dataset design and annotation to computational methods and evaluation \cite{Chaudhari2017}. The difficulties lie in capturing linguistic nuances, context-dependencies and latent meaning, due to richness of dynamic variants and figurative use of language \cite{Joshi2015}.

The automatic detection of sarcastic expressions often relies on the contrast between positive and negative sentiment \cite{Riloff2013}. This incongruence can be found on a lexical level with sentiment-bearing words, as in "\textit{I love being ignored}". In more complex linguistic settings an action or a situation can be perceived as negative, without revealing any affect-related lexical elements. The intention of the speaker as well as common knowledge or shared experience can be key aspects, as in "\textit{I love waking up at 5 am}", which can be sarcastic, but not necessarily. Similarly, verbal irony is referred to as saying the opposite of what is meant and based on sentiment contrast \cite{Grice1975}, whereas situational irony is seen as describing circumstances with unexpected consequences \cite{Lucariello1994,Shelley2001}.

Empirical studies have shown that there are specific linguistic cues and combinations of such that can serve as indicators for sarcastic and ironic expressions. Lexical and morpho-syntactic cues include exclamations and interjections, typographic markers such as all caps, quotation marks and emoticons, intensifiers and hyperboles \cite{Kunneman2015,Bharti2016}. In the case of Twitter, the usage of emojis and hashtags has also proven to help automatic irony detection.

We propose a purely character-based architecture which tackles these challenges by allowing us to use a learned representation that models features derived from morpho-syntactic cues. To do so, we use deep contextualized word representations, which have recently been used to achieve the state of the art on six NLP tasks, including sentiment analysis \cite{Peters2018}. We test our proposed architecture on 7 different irony/sarcasm datasets derived from 3 different data sources, providing state-of-the-art performance in 6 of them and otherwise offering competitive results, showing the effectiveness of our proposal. We make our code available at \url{https://github.com/epochx/elmo4irony}.

\section{Related work}

Apart from the relevance for industry applications related to sentiment analysis, sarcasm and irony detection has received great traction within the NLP research community, resulting in a variety of methods, shared tasks and benchmark datasets. Computational approaches for the classification task range from rule-based systems \cite{Riloff2013,Bharti2015} and statistical methods and machine learning algorithms such as Support Vector Machines \cite{Joshi2015,Tungthamthiti2010}, Naive Bayes and Decision Trees \cite{Reyes2013} leveraging extensive feature sets, to deep learning-based approaches. In this context, \citet{Tay2018}. delivered state-of-the-art results by using an intra-attentional component in addition to a recurrent neural network. Previous work such as the one by \citet{Veale2016} had proposed a convolutional long-short-term memory network (CNN-LSTM-DNN) that also achieved excellent results. A comprehensive survey on automatic sarcasm detection was done by \citet{Joshi2016b}, while computational irony detection was reviewed by \citet{Wallace2015}.

Further improvements both in terms of classic and deep models came as a result of the SemEval 2018 Shared Task on Irony in English Tweets \cite{VanHee2018}. The system that achieved the best results was hybrid, namely, a densely-connected BiLSTM with a multi-task learning strategy, which also makes use of features such as POS tags and lexicons \cite{Wu2018}. 

\begin{table*}[!ht]
	\centering
    \def\arraystretch{1.4}
    \footnotesize
    \begin{tabular}{l | c | c | c | c| c | c}
    \textbf{Reference} &  \textbf{Dataset} & \textbf{Train} & \textbf{Valid} & \textbf{Test} & \textbf{Total} & \textbf{Source} \\
    \hline
    \citeauthor{VanHee2018}, \citeyear{VanHee2018} & SemEval-2018 & 3,067	& 306	 & 784		& 3,834		& Twitter \\
    \citeauthor{Ptacek2014}, \citeyear{Ptacek2014} & Pt\'{a}\v{c}ek 	  & 48,007 	& 6,858  & 13,717 	& 68,582	& Twitter \\
	\citeauthor{Riloff2013}, \citeyear{Riloff2013} & Riloff 	  & 1,327	& 189	 & 381		& 1,897  	& Twitter \\
	\citeauthor{Khodak2017}, \citeyear{Khodak2017} & SARC 2.0 	  & 205,665 & 51,417 & 64,666 	& 321,748	& Reddit \\
    \citeauthor{Khodak2017}, \citeyear{Khodak2017} & SARC 2.0 pol & 10,934  & 2,734  & 3,406    & 17,074    & Reddit \\
    \citeauthor{Oraby2016}, \citeyear{Oraby2016}   & SC-V1		  & 1,396   & 199 	 & 400		& 1,995 	& Dialogues \\
    \citeauthor{Oraby2016}, \citeyear{Oraby2016}   & SC-V2		  & 3,284   & 469 	 & 939		& 4,692 	& Dialogues \\
    \end{tabular}
\caption{Benchmark datasets: Tweets, Reddit posts and online debates for sarcasm and irony detection.}
\label{table:data}
\end{table*}

\section{Proposed Approach}

The wide spectrum of linguistic cues that can serve as indicators for sarcastic and ironic expressions has been usually exploited for automatic sarcasm or irony detection by modeling them in the form of binary features in traditional machine learning.

On the other hand, deep models for irony and sarcasm detection, which are currently offer state-of-the-art performance, have exploited sequential neural networks such as LSTMs and GRUs \cite{Veale2016,zhang2016} on top of distributed word representations. Recently, in addition to using a sequential model, \citet{Tay2018} proposed to use intra-attention to compare elements in a sequence against themselves. This allowed the model to better capture word-to-word level interactions that could also be useful for detecting sarcasm, such as the \textit{incongruity} phenomenon \cite{Joshi2015}. Despite this, all models in the literature rely on word-level representations, which keeps the models from being able to easily capture some of the lexical and morpho-syntactic cues known to denote irony, such as all caps, quotation marks and emoticons, and in Twitter, also emojis and hashtags. 

The usage of a purely character-based input would allow us to directly recover and model these features. Consequently, our architecture is based on Embeddings from Language Model or ELMo \cite{Peters2018}. The ELMo layer allows to recover a rich 1,024-dimensional dense vector for each word. Using CNNs, each vector is built upon the characters that compose the underlying words. As ELMo also contains a deep bi-directional LSTM on top of this character-derived vectors, each word-level embedding contains contextual information from their surroundings. Concretely, we use a pre-trained ELMo model, obtained using the 1 Billion Word Benchmark which contains about 800M tokens of news crawl data from WMT 2011 \cite{Chelba2014}. 

Subsequently, the contextualized embeddings are passed on to a BiLSTM with 2,048 hidden units. We aggregate the LSTM hidden states using max-pooling, which in our preliminary experiments offered us better results, and feed the resulting vector to a 2-layer feed-forward network, where each layer has 512 units. The output of this is then fed to the final layer of the model, which performs the binary classification.

\section{Experimental Setup}

We test our proposed approach for binary classification on either sarcasm or irony, on seven benchmark datasets retrieved from different media sources. Below we describe each dataset, please see Table \ref{table:data} below for a summary.

\textbf{Twitter}: We use the Twitter dataset provided for the SemEval 2018 Task 3, Irony Detection in English Tweets \cite{VanHee2018}. The dataset was manually annotated using binary labels. We also use the dataset by \citet{Riloff2013}, which is manually annotated for sarcasm. Finally, we use the dataset by \citet{Ptacek2014}, who collected a user self-annotated corpus of tweets with the \textit{\#sarcasm} hashtag. 

\textbf{Reddit}: \citet{Khodak2017} collected SARC, a corpus comprising of 600.000 sarcastic comments on Reddit. We use main subset, \textit{SARC 2.0}, and the political subset, \textit{SARC 2.0 pol}.

\textbf{Online Dialogues}: We utilize the \textit{Sarcasm Corpus V1} (SC-V1) and the \textit{Sarcasm Corpus V2} (SC-V2), which are subsets of the Internet Argument Corpus (IAC). Compared to other datasets in our selection, these differ mainly in text length and structure complexity \cite{Oraby2016}. 

\begin{table*}[ht!]
  \centering
  \def\arraystretch{1.2}
  \footnotesize
  \begin{tabular}{c|c|l|c|c|c|c}
  \multicolumn{2}{c|}{\textbf{Dataset}} & \textbf{Model} & \textbf{Accuracy} & \textbf{Precision} & \textbf{Recall} & \textbf{F1-Score} \\
  \hline
  
  \multirow{12}{*}{Twitter} 
      & \multirow{4}{*}{SemEval-2018}		
      		  & \citet{Wu2018}        & \textbf{0.735}      & 0.630      & \textbf{0.801}      & \textbf{0.705}       \\ 
            & & ELMo-BiLSTM           & 0.708      & \textbf{0.696}      & 0.697      & 0.696       \\ 
            & & ELMo-BiLSTM-FULL      & 0.702      & 0.689      & 0.689      & 0.689       \\ 
            & & ELMo-BiLSTM-AUG       & 0.658      & 0.651      & 0.657      & 0.651          \\ 
      \cline{2-7}
      & \multirow{3}{*}{Riloff}		
      		  & \citet{Tay2018}       & 0.823      & 0.738      & 0.732      & 0.732       \\ 
            & & ELMo-BiLSTM			  & 0.842      & 0.759      & \textbf{0.750}      & \textbf{0.759}       \\ 
            & & ELMo-BiLSTM-FULL      & \textbf{0.858}      & \textbf{0.778 }     & 0.735      & 0.753       \\ 
            & & ELMo-BiLSTM-AUG       & 0.798      & 0.684      & 0.708      & 0.694       \\ 
      \cline{2-7}
      & \multirow{3}{*}{Pt\'{a}\v{c}ek}  		
      		  & \citet{Tay2018}       & 0.864      & 0.861      & 0.858      & 0.860       \\ 
            & & ELMo-BiLSTM			  & \textbf{0.876}      & 0.868      & 0.869      & 0.869       \\ 
            & & ELMo-BiLSTM-FULL      & 0.872      & \textbf{0.872}      & \textbf{0.872}      & \textbf{0.872}       \\ 
            & & ELMo-BiLSTM-AUG       & 0.859      & 0.859      & 0.858      & 0.859       \\ 
	 
  \hline
  \multirow{6}{*}{Dialog}  
      & \multirow{3}{*}{SC-V1}  		
      		  & \citet{Tay2018}       & 0.632      & 0.639      & 0.637      & 0.632       \\ 
            & & ELMo-BiLSTM			  & \textbf{0.646}      & \textbf{0.650}      & \textbf{0.646}      & \textbf{0.644}       \\ 
            & & ELMo-BiLSTM-FULL      & 0.633      & 0.633      & 0.633      & 0.633       \\ 
      \cline{2-7}
      & \multirow{3}{*}{SC-V2}  	
      		  & \citet{Tay2018}       & 0.729      & 0.729      & 0.729      & 0.728       \\ 
            & & ELMo-BiLSTM			  & 0.748      & 0.748      & 0.747      & 0.747        \\ 
            & & ELMo-BiLSTM-FULL      & \textbf{0.760}      & \textbf{0.760}      & \textbf{0.760 }     & \textbf{0.760}       \\ 
 
 \hline
  \multirow{6}{*}{Reddit} 
      & \multirow{3}{*}{SARC 2.0} 
      		  & \citet{Khodak2017}    & 0.758      & -          & -          & -           \\ 
            & & ELMo-BiLSTM           & \textbf{0.773}      & -          & -          & -           \\ 
            & & ELMo-BiLSTM-FULL      & 0.702      & 0.760      &   0.760    &  0.760      \\ 
  	 \cline{2-7}
      & \multirow{3}{*}{SARC 2.0 pol}
      		  & \citet{Khodak2017}    & 0.765      & -          & -          & -           \\ 
            & & ELMo-BiLSTM			  & \textbf{0.785}      & -          & -          & -           \\ 
            & & ELMo-BiLSTM-FULL      & 0.720      & 0.720      & 0.720      & 0.720        \\ 
  \end{tabular}
\caption{Summary of our obtained results.}
\label{table:results}

\end{table*}

In Table \ref{table:data}, we see a notable difference in terms of size among the Twitter datasets. Given this circumstance, and in light of the findings by \citet{VanHee2018}, we are interested in studying how the addition of external soft-annotated data impacts on the performance. Thus, in addition to the datasets introduced before, we use two corpora for augmentation purposes. The first dataset was collected using the Twitter API, targeting tweets with the hashtags \textit{\#sarcasm} or \textit{\#irony}, resulting on a total of 180,000 and 45,000 tweets respectively. On the other hand, to obtain non-sarcastic and non-ironic tweets, we relied on the SemEval 2018 Task 1 dataset \cite{Mohammad2018}. To augment each dataset with our external data, we first filter out tweets that are not in English using language guessing systems. We later extract all the hashtags in each target dataset and proceed to augment only using those external tweets that contain any of these hashtags. This allows us to, for each class, add a total of 36,835 tweets for the Pt\'{a}\v{c}ek corpus, 8,095 for the Riloff corpus and 26,168 for the SemEval-2018 corpus.

In terms of pre-processing, as in our case the preservation of morphological structures is crucial, the amount of normalization is minimal. Concretely, we forgo stemming or lemmatizing, punctuation removal and lowercasing. We limit ourselves to replacing user mentions and URLs with one generic token respectively. In the case of the SemEval-2018 dataset, an additional step was to remove the hashtags \textit{\#sarcasm}, \textit{\#irony} and \textit{\#not}, as they are the artifacts used for creating the dataset. For tokenizing, we use a variation of the Twokenizer \cite{gimpel-EtAl:2011:ACL-HLT2011} to better deal with emojis.

Our models are trained using Adam with a learning rate of 0.001 and a decay rate of 0.5 when there is no improvement on the accuracy on the validation set, which we use to select the best models. We also experimented using a slanted triangular learning rate scheme, which was shown by \citet{jeremyruder2018} to deliver excellent results on several tasks, but in practice we did not obtain significant differences. We experimented with batch sizes of 16, 32 and 64, and dropouts ranging from 0.1 to 0.5. The size of the LSTM hidden layer was fixed to 1,024, based on our preliminary experiments. We do not train the ELMo embeddings, but allow their dropouts to be active during training.

\section{Results}

Table \ref{table:results} summarizes our results. For each dataset, the top row denotes our baseline and the second row shows our best comparable model. Rows with FULL models denote our best single model trained with all the development available data, without any other preprocessing other than mentioned in the previous section. In the case of the Twitter datasets, rows indicated as AUG refer to our the models trained using the augmented version of the corresponding datasets.

For the case of the SemEval-2018 dataset we use the best performing model from the Shared Task as a baseline, taken from the task description paper \cite{VanHee2018}. As the winning system is a voting-based ensemble of 10 models, for comparison, we report results using an equivalent setting. For the Riloff, Pt\'{a}\v{c}ek, SC-V1 and SC-V2 datasets, our baseline models are taken directly from \citet{Tay2018}. As their pre-processing includes truncating sentence lengths at 40 and 80 tokens for the Twitter and Dialog datasets respectively, while always removing examples with less than 5 tokens, we replicate those steps and report our results under these settings. Finally, for the Reddit datasets, our baselines are taken from \citet{Khodak2017}. Although their models are trained for binary classification, instead of reporting the performance in terms of standard classification evaluation metrics, their proposed evaluation task is predicting which of two given statements that share the same context is sarcastic, with performance measured solely by accuracy. We follow this and report our results.

In summary, we see our introduced models are able to outperform all previously proposed methods for all metrics, except for the SemEval-2018 best system. Although our approach yields higher Precision, it is not able to reach the given Recall and F1-Score. We note that in terms of single-model architectures, our setting offers increased performance compared to \citet{Wu2018} and their obtained F1-score of 0.674. Moreover, our system does so without requiring external features or multi-task learning. For the other tasks we are able to outperform \citet{Tay2018} without requiring any kind of intra-attention. This shows the effectiveness of using pre-trained character-based word representations, that allow us to recover many of the morpho-syntactic cues that tend to denote irony and sarcasm.

Finally, our experiments showed that enlarging existing Twitter datasets by adding external soft-labeled data from the same media source does not yield improvements in the overall performance. This complies with the observations made by \citet{VanHee2018}. Since we have designed our augmentation tactics to maximize the overlap in terms of topic, we believe the soft-annotated nature of the additional data we have used is the reason that keeps the model from improving further.

\section{Conclusions}

We have presented a deep learning model based on character-level word representations obtained from ELMo. It is able to obtain the state of the art in sarcasm and irony detection in 6 out of 7 datasets derived from 3 different data sources. Our results also showed that the model does not benefit from using additional soft-labeled data in any of the three tested Twitter datasets, showing that manually-annotated data may be needed in order to improve the performance in this way.

\bibliography{Sarcasm}
\bibliographystyle{acl_natbib_nourl}
\end{document}